\documentclass{isprs} 
\usepackage{setspace}
\usepackage{geometry} 
\usepackage{epstopdf}
\usepackage[labelsep=period]{caption}  
\usepackage{hyphenat}
\usepackage{breakcites} 

\usepackage[utf8]{inputenc} 
\usepackage[T1]{fontenc}    
\usepackage{microtype}

\usepackage{siunitx}
\usepackage{etoolbox}
\robustify\bfseries
\usepackage{upgreek}
\usepackage[hidelinks]{hyperref}
\usepackage{tabularx}
\usepackage{booktabs}
\usepackage{subfig}
\usepackage{multirow}
\usepackage{color}
\usepackage{natbib}
\usepackage{paralist}
\usepackage{amsmath}
\usepackage{amssymb}
\usepackage[super]{nth}
\usepackage[nameinlink]{cleveref}
\usepackage{float}
\usepackage{stfloats}
\usepackage{soul}

\begin{document}

\begin{figure*}
\centering
    \begin{minipage}{0.7\textwidth}
        \Huge
        Update \\ \\ 
        \normalsize
        An extended and peer-reviewed version of this paper exists.\\
        Please read and possibly cite the newer version:\\ \\
        Riese, F.M., Keller, S. and Hinz, S., 2020. Supervised and Semi-Supervised Self-Organizing Maps for Regression and Classification Focusing on Hyperspectral Data. \textit{Remote Sensing} 12(1). \href{https://doi.org/10.3390/rs12010007}{doi:10.3390/rs12010007} \\ \\
        $\boldsymbol\rightarrow$ \href{https://doi.org/10.3390/rs12010007}{\textcolor{red}{\textbf{Link to the new paper}}}
    \end{minipage}
\end{figure*}

\newpage

\title{SuSi: Supervised Self-Organizing Maps for Regression and Classification in Python}

\author{
Felix~M. Riese\textsuperscript{1}, Sina Keller\textsuperscript{1}}
\address{
	\textsuperscript{1 }Institute of Photogrammetry and Remote Sensing,
	\\ Karlsruhe Institute of Technology, Karlsruhe, Germany 
	\\(felix.riese, sina.keller)@kit.edu\\
}

\commission{}{}
\workinggroup{}
\icwg{}

\abstract{
In many research fields, the sizes of the existing datasets vary widely.
Hence, there is a need for machine learning techniques which are well-suited for these different datasets.
One possible technique is the self-organizing map (SOM), a type of artificial neural network which is, so far, weakly represented in the field of machine learning.
The SOM's unique characteristic is the neighborhood relationship of the output neurons.
This relationship improves the ability of generalization on small datasets.
SOMs are mostly applied in unsupervised learning and few studies focus on using SOMs as supervised learning approach.
Furthermore, no appropriate SOM package is available with respect to machine learning standards and in the widely used programming language Python.
In this paper, we introduce the freely available \textbf{Su}pervised \textbf{S}elf-organ\textbf{i}zing maps (SuSi) Python package which performs supervised regression and classification.
The implementation of SuSi is described with respect to the underlying mathematics.
Then, we present first evaluations of the SOM for regression and classification datasets from two different domains of geospatial image analysis.
Despite the early stage of its development, the SuSi framework performs well and is characterized by only small performance differences between the training and the test datasets.
A comparison of the SuSi framework with existing Python and \textsf{R} packages demonstrates the importance of the SuSi framework.
In future work, the SuSi framework will be extended, optimized and upgraded e.g. with tools to better understand and visualize the input data as well as the handling of missing and incomplete data.
}

\keywords{Self-Organizing Maps, Machine Learning, Unsupervised Learning, Supervised Learning, Python}

\maketitle

\section{Introduction}
\label{sec:intro}

With increasing computing power and increasing amount of data, artificial neural networks (ANN) have become a standard tool for regression and classification tasks.
Feed-forward neural networks and convolutional neural networks (CNN) are the most common types of ANN in current research.
According to the \textit{no free lunch theorem} by~\cite{Wolpert95nofree}, a variety of possible tools is necessary to be able to adapt to new tasks.

One underrepresented type of ANNs is the self-organizing map (SOM).
The SOM was introduced by \cite{kohonen1982self, kohonen1990the, kohonen1995selforganizing, kohonen2013essentials}.
It is a shallow ANN architecture consisting of an input layer and a 2-dimensional (2D) grid as output layer.
The latter is fully connected to the input layer.
Besides, the neurons on the output grid are interconnected to each other through a neighborhood relationship.
Changes on the weights of one output neuron also affect the neurons in its neighborhood.
This unique characteristic decreases overfitting of the training datasets.
Further, the 2D output grid visualizes the results of the SOM comprehensibly.
This plain visualization does not exist in the majority of ANNs.

In the following, we give a brief overview of the various SOM applications in different fields of research.
Most SOMs are applied in unsupervised learning like clustering, visualization and dimensionality reduction.
A good overview of SOMs as unsupervised learners and their applications in the research field of water resources is presented by \cite{kalteh2008review}.
SOMs are also applied to the scientific field of maritime environment research \citep{lobo2009self}.
One major application of SOMs is clustering data \citep{vesanto2000clustering} and cluster-wise regression \citep{muruzbal2012}.

SOMs can be combined with other machine learning techniques.
For example, the output of the unsupervised SOM is used by \cite{hsu2009a} as input for a support vector regressor to forecast stock prices.
\cite{hsu2002selforganizing} present the combination of SOMs with linear regression in hydrology.
SOMs can also be used for data fusion, e.g. for plant disease detection \citep{moshou2005plant}.
\cite{hagenbuchner2005a} add majority voting to SOMs for the application as supervised classifier.
The combination of SOMs and nearest-neighbor classification is shown by \citet{ji2006land} for the classification of land use.
Additional combinations of unsupervised SOMs and supervised algorithms used for classification are presented by \cite{martinez2002hyperspectral, zaccarelli2003spectral, zhong2006an, fessant2001comparison}.
One example for the application of SOMs to solve non-linear regression tasks is presented by \cite{hecht2015using} in the field of robotics.

Two of the most popular programming languages for machine learning applications are Python and \textsf{R}.
Programming frameworks like scikit-learn~\citep{scikit-learn} in Python have simplified the application of existing machine learning techniques considerably.
While in the programming language \textsf{R} the \textit{kohonen} package \citep{wehrens2018flexible} provides a standardized framework for SOMs, in Python there exists no such standard SOM package, yet.

In this paper, we introduce the Python package \textbf{Su}pervised {\textbf{S}elf}- {organ\textbf{i}zing} maps (SuSi) framework for regression and classification.
It is the first Python package that provides unsupervised and supervised SOM algorithms for easy usage.
The SuSi framework is available freely on GitHub \citep{riese2019susi}.
This ensures the maintainability and the opportunity for future upgrades by the authors as well as any member of the community.
The implementation was briefly introduced in~\cite{riese2018introducing} with respect to the regression of soil-moisture~\citep{keller2018developing} and the estimation of water quality parameters~\citep{keller2018hyperspectral}.
The main contributions of this paper are:
\begin{compactitem}
	\item the implementation of the SuSi framework including the combination of an unsupervised SOM and a supervised SOM for regression and classification tasks that is able to perform on small as well as on large datasets without significant overfitting,
	\item the mathematical description of all implemented processes of the SuSi framework with a consistent variable naming convention for the unsupervised SOM in \Cref{sec:unsuper} and the supervised SOM in \Cref{sec:super},
	\item a first evaluation of the regression and classification capabilities of the SuSi framework in \Cref{sec:eval:sub:reg,sec:eval:sub:cla},
	\item a detailed comparison of the SuSi framework with existing Python and \textsf{R} packages based on a list of requirements in \Cref{sec:eval:sub:comp} and
	\item an outlook into the future of the SuSi framework in \Cref{sec:conclusion}.
\end{compactitem}

\section{SuSi part 1: unsupervised learning}
\label{sec:unsuper}

\begin{figure}[p]
	\centering
	\includegraphics[width=0.35\textwidth]{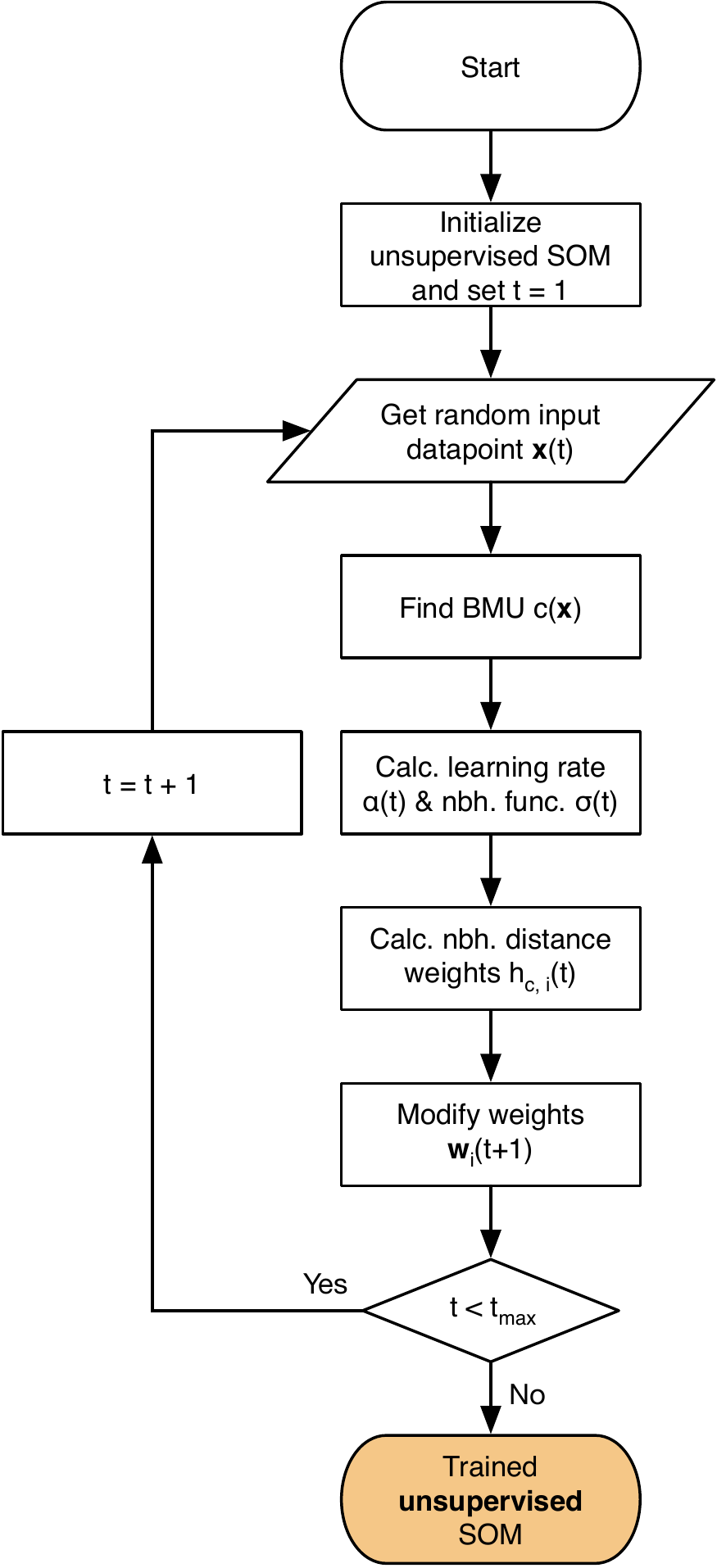}
	\caption{Flowchart of the unsupervised SOM algorithm resulting in the trained unsupervised SOM (orange).}
	\label{fig:flow_som_unsuper}
\end{figure}

In the following, we describe the architecture and mathematics of the unsupervised part of the SuSi framework.
The grid of a SOM, the map, can be implemented in different topologies.
In this paper, we use the simplest topology: the 2D rectangular grid consisting of $n_{\text{row}}\times n_{\text{column}}$ nodes.
This grid is fully connected to the input layer as each node is connected to all $n$ input features via $n$ weights.
The variable naming conventions of the SuSi framework are given in \Cref{tab:conventions}.
The training process of the unsupervised SOM is illustrated in \Cref{fig:flow_som_unsuper} and consists of the following steps:
\begin{compactenum}
	\item \label{en:un_init} Initialize the SOM.
	\item \label{en:un_random} Get random input datapoint.
	\item \label{en:un_bmu} Find best matching unit (BMU) (cf. \Cref{sec:unsuper:sub:bmu}).
	\item \label{en:un_lrnbhf} Calculate learning rate (cf. \Cref{sec:unsuper:sub:learningrate}) and neighborhood function (cf. \Cref{sec:unsuper:sub:nbh}).
	\item \label{en:un_nbhdw} Calculate neighborhood distance weight matrix (cf. \Cref{sec:unsuper:sub:nbhdist}).
	\item \label{en:un_mod} Modify SOM weight matrix (cf. \Cref{sec:unsuper:sub:adapt}).
	\item Repeat from step~\ref{en:un_random} until the maximum number of iterations is reached.
\end{compactenum}

\begin{table}[p]
	\centering
	\caption{Variable naming conventions of the SuSi framework.}
	\begin{tabular}{ll}
		\toprule
		Variable & Description\\
		\midrule
		$n$ & Number of features of a datapoint\\
		$N$ & Number of datapoints\\
		$n_{\text{row}}$ & Number of rows on the SOM grid\\
		$n_{\text{column}}$ & Number of columns on the SOM grid\\
		$t$ & Number of current iteration\\
		$t_{\max}$ & Number of maximum iterations, $t < t_{\max}$\\
		$\boldsymbol{x}(t)$ & Datapoint at iteration $t$ with $\boldsymbol{x} \in\mathbb{R}^n$\\
		$y(t)$ & Label of datapoint $\boldsymbol{x}(t)$\\
		$c(\boldsymbol{x})$ & Best matching unit (BMU) of datapoint $\boldsymbol{x}(t)$\\
		& with $c\in\mathbb{R}^2$\\
		$\alpha(t)$ & Function of the learning rate\\
		$\alpha_0$ & Start value of the learning rate\\
		$\sigma(t)$ & Neighborhood function\\
		$\sigma_0$ & Start value of the neighborhood function\\
		$\sigma_{\text{end}}$ & End value of the neighborhood function\\
		$h_{c,i}$ & Neighborhood distance weight between\\
		& BMU $c$ and SOM node $i$\\
		$\boldsymbol{w}_i(t)$ & Weight of node $i$ at iteration $t$ with $\boldsymbol{w}_i \in\mathbb{R}^n$\\
		\bottomrule
	\end{tabular}
	\label{tab:conventions}
\end{table}

The initialization approach of a SOM mainly affects the speed of its training phase.
The SOM weights of the SuSi framework are initialized randomly at this stage of development.
\cite{attik2005artificial} and \cite{akinduko2016som}, for example, propose more sophisticated initialization approaches like applying a principal component analysis.
In the following subsections, the training of an unsupervised SOM is described in detail.

\subsection{Finding the best matching unit}
\label{sec:unsuper:sub:bmu}

During the search for the best matching unit (BMU), the input datapoint is compared to all weights on the SOM grid.
The SOM node that is the closest one to the input node according to the chosen distance metric is the BMU.
Several distance metrics can be applied.
The most common distance metric is the \textit{Euclidean distance} defined as
\begin{align}
	d(\boldsymbol{x}, \boldsymbol{x}') = \sqrt{\sum_{i=1}^n (x_i-x_i')^2},\label{eq:disteuclid}
\end{align}
with a dimension $n$ of the vectors $\boldsymbol{x}, \boldsymbol{x}'$.
Another possible choice is the \textit{Manhattan distance} which is defined as the sum of the absolute distances per element:
\begin{align}
	d(\boldsymbol{x}, \boldsymbol{x}') = \sum_{i=1}^n |x_i-x_i'|.\label{eq:distmanh}
\end{align}
The \textit{Tanimoto distance} as third option is defined as the distance or dissimilarity between two boolean (binary: $T, F$) vectors $\boldsymbol{x}, \boldsymbol{x}'$:
\begin{align}
	d(\boldsymbol{x}, \boldsymbol{x}') = \frac{R}{c_{TT} + c_{FF}+R} \qquad \text{with } R=2(c_{TF} + c_{FT}), \label{eq:disttanimoto}
\end{align}
with $c_{ij}$ as the number of occurrences of $x_k=i$ and $x_k'=j$ for all elements $k$ as defined in~\cite{scipy2001jones}.
The \textit{Mahalanobis distance} between two 1-dimensional vectors $\boldsymbol{x}, \boldsymbol{x}'$ is defined as
\begin{align}
	d(\boldsymbol{x}, \boldsymbol{x}') = \sqrt{(\boldsymbol{x}-\boldsymbol{x}')V^{-1}(\boldsymbol{x}-\boldsymbol{x}')^T},
\end{align}
with the covariance matrix $V$ of the two vectors.
The default distance metric of the SuSi framework is the Euclidean distance defined in \Cref{eq:disteuclid}.

\subsection{Learning rate}
\label{sec:unsuper:sub:learningrate}

For a faster convergence and to prevent oscillations, decreasing learning rates are often implemented in ANNs.
The learning rate $\alpha$ of the SOM training is a function that decreases from a value $\alpha_0$ with increasing number of iterations.
In general, there is an infinite number of possible functions for the learning rate.
In the following, we present several functions implemented into the SuSi framework.
In \cite{natita2016appropriate}, different learning rates for SOMs are introduced:

\begin{align}
	\alpha(t) &= \alpha_0\cdot\frac{1}{t}\label{eq:lr_inv},\\
	\alpha(t) &= \alpha_0\cdot\left( 1 - \frac{t}{t_{\max}}\right)\label{eq:lr_linear},\\
	\alpha(t) &= \alpha_0^{t/t_{\max}} \qquad \text{with } \alpha_0 = 0.005\label{eq:lr_root}.
\end{align}

\begin{figure}
	\centering
	\includegraphics[width=0.45\textwidth]{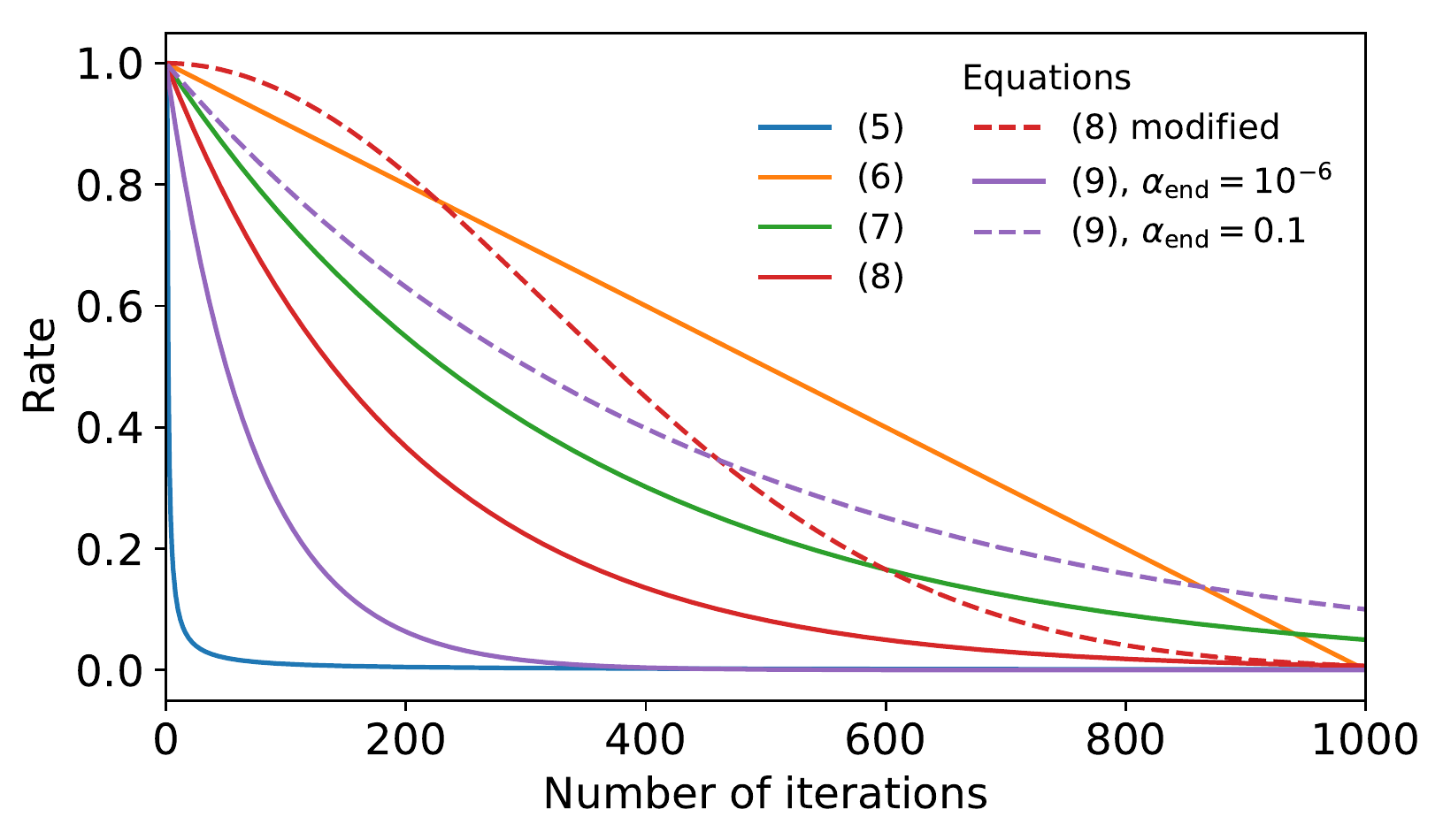}
	\caption{Comparison of different choices for the functional behavior of decreasing rates, which are implemented as learning rate and neighborhood function, on the number of iterations $t$ ($t_{\text{max}}=1000, \alpha_{0}=1$).}
	\label{fig:learningrate}
\end{figure}

In~\cite{S2012RecurrentSM}, the following learning rate was applied:
\begin{align}
	\alpha(t) = \alpha_0\cdot \exp(-t/t_{\max})\label{eq:lr_e}.
\end{align}
The implementation of~\cite{barreto2004identification} includes not only a start value for the learning rate but also an end value $\alpha_{\text{end}}$:
\begin{align}
	\alpha(t) = \alpha_0\cdot \left(\frac{\alpha_{\text{end}}}{\alpha_{0}}\right)^{t/t_{\max}}.\label{eq:lr_end}
\end{align}
In \Cref{fig:learningrate}, some examples for the behaviour of the functions are plotted.
The default learning rate function of the SuSi framework is set according to \Cref{eq:lr_end}.

\subsection{Neighborhood function}
\label{sec:unsuper:sub:nbh}

Similar to the learning rate, the neighborhood function is monotonically decreasing.
We include the following three widely-used functions in the SuSi framework.
Equivalent to \Cref{eq:lr_linear}, the neighborhood function in~\cite{matsushita2010batchlearning} is defined as
\begin{align}
	\sigma(t) = \sigma_0\cdot\left(1-\frac{t}{t_{\text{max}}}\right),\label{eq:nbhfct_inv}
\end{align}
with $\sigma_0$ as initial value of the neighborhood function.
In~\cite{S2012RecurrentSM}, the neighborhood function is implemented as
\begin{align}
	\sigma(t) = \sigma_0\cdot\exp(-t/t_{\max})\label{eq:nbhfct_e},
\end{align}
equivalent to \Cref{eq:lr_e}.
In~\cite{barreto2004identification}, the neighborhood function is defined similarly to \Cref{eq:lr_end} as
\begin{align}
	\sigma(t) = \sigma_0\cdot \left(\frac{\sigma_{\text{end}}}{\sigma_{0}}\right)^{t/t_{\max}}.\label{eq:nbhfct_end}
\end{align}
The default neighborhood function of the SuSi framework is set according to \Cref{eq:nbhfct_inv}.

\subsection{Neighborhood distance weight}
\label{sec:unsuper:sub:nbhdist}

The neighborhood distance weight is a function of the number of iterations and the distance $d(c, i)$ between the BMU $c$ and every other node $i$ on the SOM grid.
The distance $d(c, i)$ between the BMU $c$ and node $i$ is defined as the Euclidean distance (cf. \Cref{eq:disteuclid}) on the map grid.
In this paper, we give two examples for neighborhood distance weights.
\cite{matsushita2010batchlearning} use a \textit{Pseudo-Gaussian} neighborhood distance weight.
The weight between the BMU $c$ and the node $i$ on the SOM grid is defined as
\begin{align}
	h_{c,i}(t) = \exp\left( -\frac{d^2}{2\cdot \sigma(t)^2} \right)\label{eq:nbhdwgauss},
\end{align}
with the neighborhood function from \Cref{eq:nbhfct_inv} and the Euclidean distance $d(c, i)$ on the SOM grid.
This definition of a neighborhood distance weight is the default setting of the SuSi framework.
Another possible neighborhood distance weight is the \textit{Mexican Hat}~\citep{kohonen1995selforganizing} defined as
\begin{align}
	h_{c,i}(t) = \left( 1 - \frac{d^2}{\sigma^2} \right)\exp\left( -\frac{d^2}{2\cdot \sigma(t)^2} \right),\label{eq:nbhdwmexican}
\end{align}
again with neighborhood function of \Cref{eq:nbhfct_inv} and Euclidean distance $d(c, i)$ on the SOM grid.
The implications of the chosen neighborhood distance weight definitions on the SOM are investigated in e.g.~\cite{ritter1992neural} and \cite{horowitz1995convergence}.

\subsection{Adapting weights}
\label{sec:unsuper:sub:adapt}

The two most commonly used approaches to adapt the SOM weights are the \textit{online} and the \textit{batch} mode.
The weights of the SOM are adapted based on the learning rate and the neighborhood distance weight.
The \textit{online mode} is described in detail in~\cite{kohonen2013essentials}.
After each iteration, all weights of the SOM are adapted to the current datapoint $\boldsymbol{x}(t)$ as follows:
\begin{align}
	\boldsymbol{w}_i(t+1) = \boldsymbol{w}_i(t) + \alpha(t)\cdot h_{c,i}(t)\cdot\left( \boldsymbol{x}(t) - \boldsymbol{w}_i(t)\right),\label{eq:som_online}
\end{align}
with neighborhood function $h_{c,i}(t)$, learning rate $\alpha(t)$, weight vector $\boldsymbol{w}_i(t)$ of node $i$ at iteration $t$.
In the \textit{batch mode}~\citep{kohonen2002how, matsushita2010batchlearning}, the whole dataset consisting of $N$ datapoints is used in every iteration.
Each weight $\boldsymbol{w}_i$ is adapted as follows:
\begin{align}
	\boldsymbol{w}_i(t+1) = \frac{\sum_{j=1}^N h_{c,i}(t) \cdot\boldsymbol{x}_j}{\sum_{j=1}^N h_{c,i}(t) },\label{eq:som_batch}
\end{align}
with the neighborhood function $h_{c, i}(t)$ from \Cref{eq:nbhdwgauss} and the weight vector $\boldsymbol{w}_i(t+1)$ of node $i$ at iteration $t+1$.
In this first stage of development, the SuSi package provides only the online algorithm.

\subsection{Trained unsupervised SOM}
\label{sec:unsuper:sub:trained}

After reaching the maximum number of iterations $t_{\text{max}}$, the unsupervised SOM is fully trained.
New datapoints can be allocated to their respective BMU which will be used in \Cref{sec:super}.
Note that not every node on the SOM grid has to be linked to a datapoint from the training dataset, since there can be more SOM nodes than datapoints.

\section{SuSi part 2: supervised learning}
\label{sec:super}

To apply the SuSi framework for solving supervised regression or classification tasks, we attach a second SOM to the unsupervised SOM.
The flowchart of the second, supervised SOM is illustrated in \Cref{fig:flow_som_super}.
The two SOMs differ with respect to the dimension of the weights and their estimation algorithm.
The weights of the unsupervised SOM have the same dimension as the input data.
Thus, adapting these weights often changes the BMU for each input datapoint. 
In contrast, the weights of the supervised SOM have the same dimension as the target variable of the respective task.
One has to distinguish between two cases: regression and classification.
In the regression case, the weights are one-dimensional and contain a continuous number.
In the classification case, the weights contain a class.
By combining  the unsupervised and the supervised SOM, the former is used to select the BMU for each datapoint while the latter links the selected BMU to a specific estimation.
In the following, we describe the different implementations for regression and classification tasks.

\subsection{Regression}
\label{sec:super:sub:reg}

The implementation of the regression SOM is described in~\cite{riese2018introducing} using the example of the soil-moisture regression based on hyperspectral data.
The training of the regression SOM proceeds analogous to the unsupervised SOM: first, the SOM is initialized randomly.
Again, it iterates randomly through the dataset (cf. Step~\ref{en:un_init}).
In each iteration, the BMU is found for the current datapoint based on the trained unsupervised SOM (cf. Steps~\ref{en:un_random}, \ref{en:un_bmu}).
The BMUs do not change for the datapoints during the training since the unsupervised SOM is fully trained.
Then, the neighborhood function, the learning rate and the neighborhood distance weight matrix are calculated similarly to the algorithm of the unsupervised SOM (cf. Steps~\ref{en:un_lrnbhf}, \ref{en:un_nbhdw}).
Finally, the weights are adapted to the label $y(t)$ of the input datapoint $\boldsymbol{x}(t)$ (cf. Step~\ref{en:un_mod}).

\begin{figure}[H]
	\centering
	\includegraphics[width=0.45\textwidth]{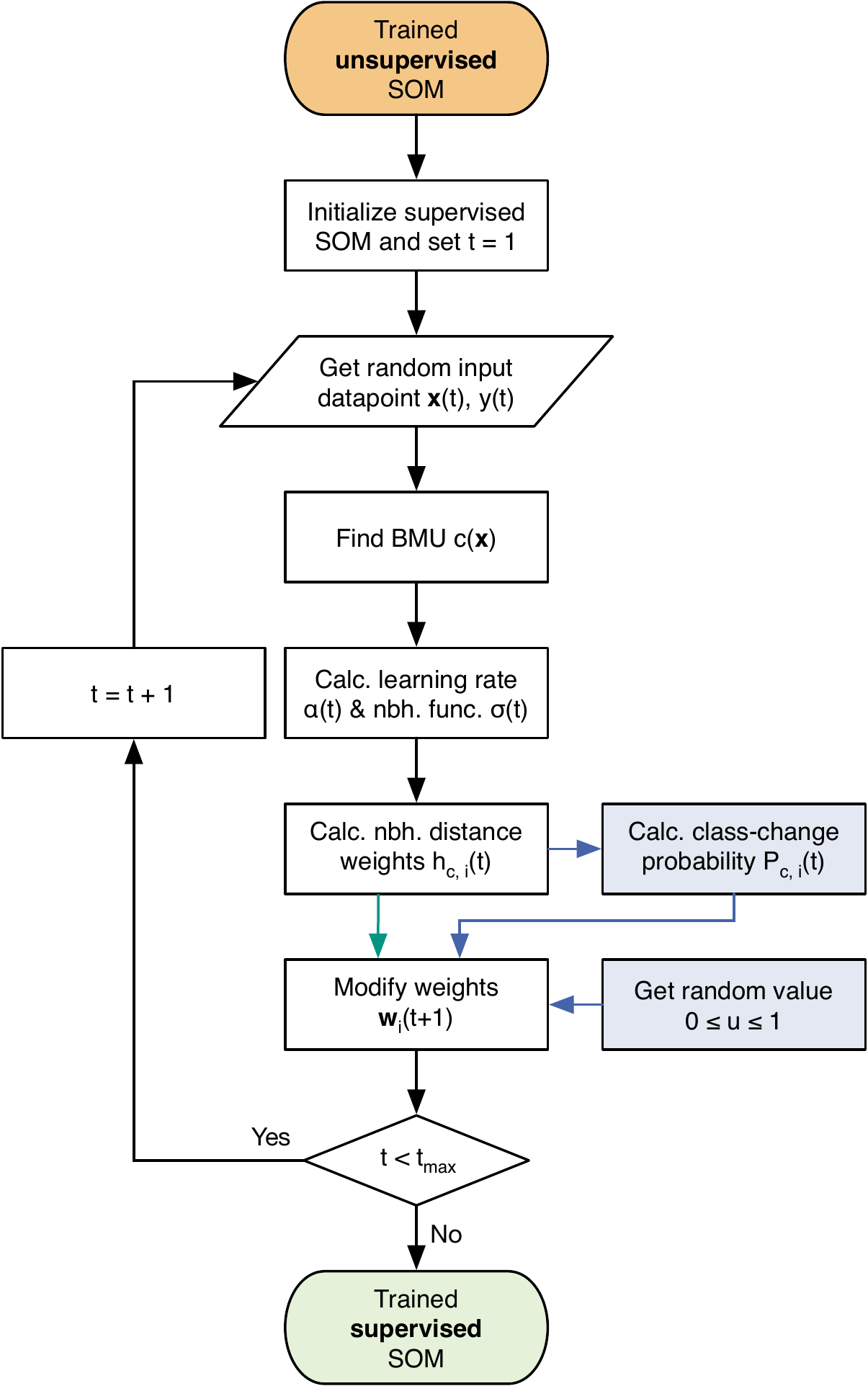}
	\caption{Flowchart of the algorithms for the regression SOM (black and cyan) and the classification SOM (black and blue). The "Trained unsupervised SOM" (orange) is the result of the unsupervised SOM algorithm illustrated in \Cref{fig:flow_som_unsuper}.}
	\label{fig:flow_som_super}
\end{figure}

In the case of the regression SOM, the label is a continuous value and the weights of the regression SOM can be modified similarly to the process described in \Cref{sec:unsuper:sub:adapt}.
After the training (and in the case of a 1-dimensional target variable), the regression SOM consists of a map with a continuous distribution of the regression target variable.
To apply the trained regression SOM to a new dataset, the BMUs needs to be found by the unsupervised SOM.
For each datapoint in the new dataset, the estimated output value of the SuSi framework is the weight of the found BMU on the regression SOM.
The regression SOM is illustrated in \Cref{fig:flow_som_super}.

\subsection{Classification}
\label{sec:super:sub:cla}

In the case of a classification task, the labels are discrete.
In contrast to the commonly used majority voting approach \citep[cf.][]{hagenbuchner2005a}, we have implemented a training process similar to the adaptation approach of the unsupervised SOM (cf. \Cref{sec:unsuper:sub:adapt}):
\begin{compactenum}
	\item \label{en:cla_init} Initialize the classification SOM.
	\item \label{en:cla_random} Get random input datapoint with label.
	\item Find BMU based on trained unsupervised SOM.
	\item Calculate learning rate and neighborhood function.
	\item \label{en:cla_nbhdw} Calculate neighborhood distance weight.
	\item \label{en:cla_prob} Calculate class-change probability matrix.
	\item \label{en:cla_mod} Modify classification SOM weight matrix.
	\item Repeat from step~\ref{en:cla_random} until the maximum number of iterations is reached.
\end{compactenum}

The classification SOM is illustrated in \Cref{fig:flow_som_super}.
The initialization in step~\ref{en:cla_init} contains a simple majority vote: each node is assigned to the class representing the majority of datapoints allocated to the respective node.
The steps~\ref{en:cla_random} to~\ref{en:cla_nbhdw} are implemented similarly to the regression SOM in \Cref{sec:super:sub:reg}.
To modify the discrete weights of the classification SOM, we introduce the class-change probability $P_{c, i}(t)$ in step~\ref{en:cla_prob}.
In the regression SOM, the SOM nodes around the BMU are adapted to the current datapoint with a certain probability depending on the learning rate and the neighborhood distance weight.
Since the labels $y$ are discrete in a classification task, this process needs to be adapted.
In the following, we explain our proposed adaptation.

For datasets with imbalanced class distributions, meaning datasets with significantly different number of datapoints per class, we provide the possibility to re-weight the dataset.
The optional class weight is defined as 
\begin{align}
	w_{\text{class}(j)} = \begin{cases}
		N / (n_{\text{classes}}\cdot N_j) & \text{if class weighting}\\
		1 & \text{otherwise}
	\end{cases},\label{eq:classweights}
\end{align}
with the number of datapoints $N$, the number of datapoints $N_j$ of class $j$ and the number of classes $n_{\text{classes}}$.
Similar to \Cref{eq:som_online}, we define a term that affects the modifying of the SOM weights.
Since the modifications need to be discrete, we work with probabilities.
The probability for a class change of node $i$ with BMU $c(\boldsymbol{x})$ of the datapoint $\boldsymbol{x}(t)$ with label $y(t)$ is defined as
\begin{align}
	P_{c, i}(t) = w_{y(t)}\cdot \alpha(t) \cdot h_{c, i}(t)\label{eq:classproba},
\end{align}
with the class weight $w_{y(t)}$ (cf. \Cref{eq:classweights}), the learning rate $\alpha(t)$ (cf. \Cref{sec:unsuper:sub:learningrate}) and the neighborhood distance weight $h_{c, i}(t)$ (cf. \Cref{sec:unsuper:sub:nbhdist}).
To decide if a node changes its assigned class, a binary decision rule is created based on this probability.
A simple threshold of e.g. \num{0.5} would lead to a static SOM after a certain number of iterations.
Therefore, we include randomization into the decision process.
For every node in every iteration, a random number $u_i(t)$ is generated which is uniformly distributed between \num{0} and \num{1}.
The modification of the weights is then implemented based on the class change probability $P_{c, i}(t)$ defined in \Cref{eq:classproba} as follows:
\begin{align}
	w_i(t+1) = 
	\begin{cases}
		y(t) & \quad\text{if } u_i(t) < P_{c, i}(t)\\
		w_i(t) & \quad\text{otherwise}
	\end{cases},\label{eq:som_class}
\end{align}
with the label $y(t)$ linked the datapoint $\boldsymbol{x}(t)$ of the current iteration $t$.
After the maximum number of iterations is reached, the classification SOM is fully trained.
Then, every node on the SOM grid is assigned to one class of the dataset.
To apply the classification SOM on new data, the BMU needs to be found for each datapoint with the unsupervised SOM.
This process is similar to the one in the trained regression SOM.
The estimation of the classification SOM for this datapoint is equivalent to the weight of the neuron in the classification SOM at the position of the selected BMU.

\section{Evaluation}
\label{sec:eval}

For a first evaluation of the regression and classification capabilities of the introduced SuSi framework, we rely on two datasets from different domains of geospatial image analysis.
The regression is evaluated in \Cref{sec:eval:sub:reg} with a hyperspectral dataset on the target variable soil moisture.
The evaluation of the classification SOM in \Cref{sec:eval:sub:cla} is performed based on the freely available Salinas valley dataset for land cover classification from hyperspectral data.
The results of the two different SOM applications are compared against a random forest (RF) estimator~\citep{breiman2001random}.
Finally, the SuSi package is compared to existing SOM packages in the programming languages Python and \textsf{R} in \Cref{sec:eval:sub:comp}.

\subsection{Regression of soil moisture}
\label{sec:eval:sub:reg}
%

The performance of the regression SOM is evaluated based on the soil-moisture dataset measured during a field campaign and published in~\cite{riesekeller2018}.
A similar evaluation is published in~\cite{riese2018introducing} with a preceding version of the SOM algorithm and code.
The dataset consists of 679 datapoints collected by a Cubert UHD 285 camera.
One datapoint consists of 125 bands between \SIrange{450}{950}{\nano\meter}.
A soil moisture sensor measured the reference values in a range of \SIrange{25}{42}{\percent} soil moisture.
For the validation of the estimation performance and the generalization capabilities of the SOM, the dataset is randomly divided into a training and a test subset in the ratio $1\mathbin{:}1$.
The training of the estimator is performed on the training subset and the evaluation is performed on the test subset.

The regression SOM is set up with default parameters with the exception of the grid size and the number of iterations.
The grid size of the SOM is $35\times 35$.
The unsupervised and the supervised SOMs are trained with $2500$ iterations each.
These hyperparameters can be further optimized depending on the applied dataset.
The RF regressor is set up with \num{1000} estimators and the scikit-learn default hyperparameters~\citep[cf.][]{scikit-learn}.
For the evaluation, we choose the coefficient of determination $R^2$.

The regression SOM achieves $R^2 = \SI{90.2}{\percent}$ on the test subset.
This score implies that the regression SOM is able to generalize on this dataset.
Interestingly, the result for the training subset is only marginally better with  $R^2_{\text{train}} = \SI{90.7}{\percent}$.
In comparison, the RF regressor results in $R^2 = \SI{93.6}{\percent}$ and $R^2_{\text{train}} = \SI{99.2}{\percent}$ on the dataset.
To conclude, the SOM seems to be robust regarding overfitting.
In this case, the $R^2_{\text{train}}$ score could function as \textit{out-of-bag estimate} for the dataset similar to~\cite{breiman2001random}.
When dealing with small datasets, the SOM provides the advantage of not necessitating on the split of the dataset.

\begin{figure}[tb]
	\centering
	\subfloat[]{\includegraphics[height=4.2cm, keepaspectratio]{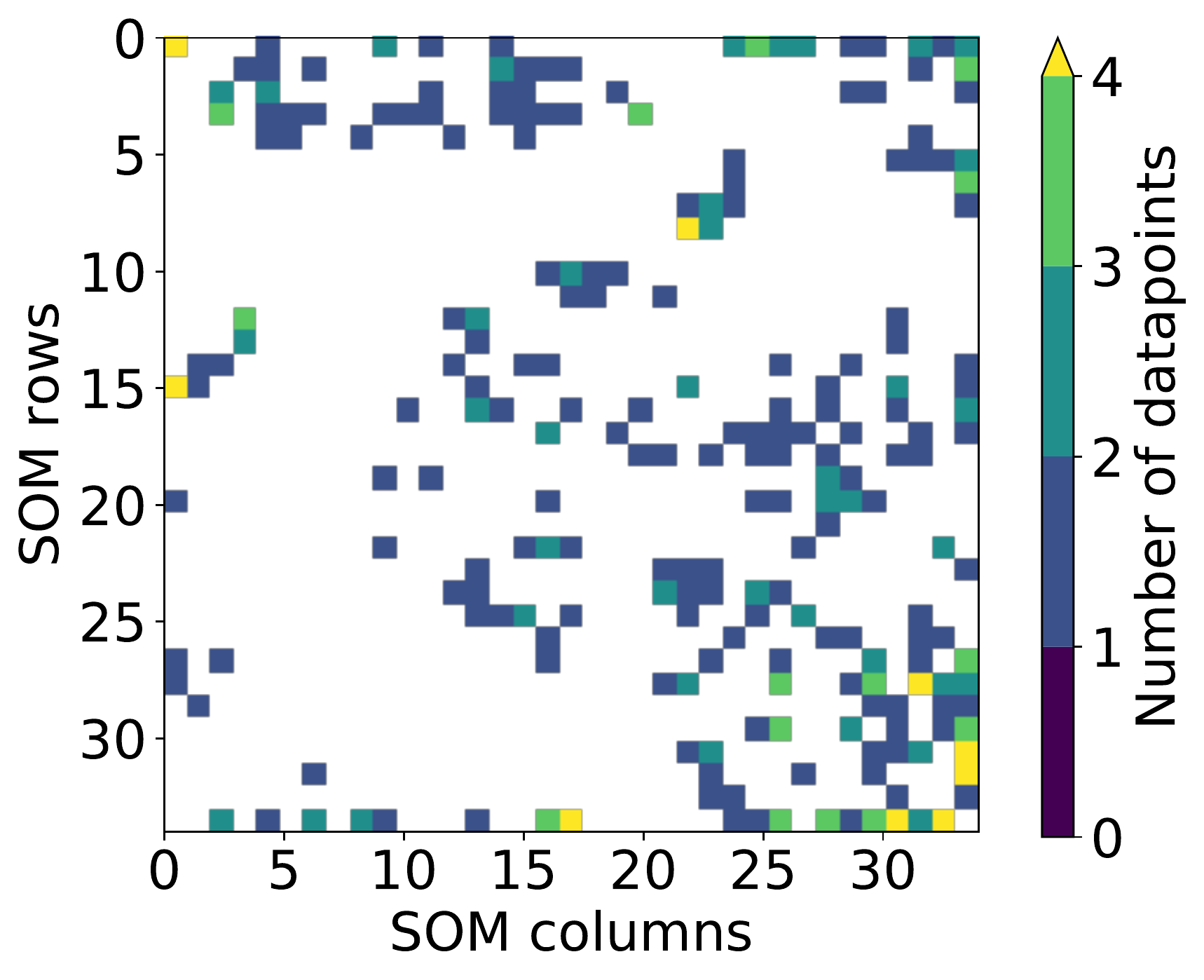}\label{fig:eval_reg_histo}}\
	\subfloat[]{\includegraphics[height=4.2cm, keepaspectratio]{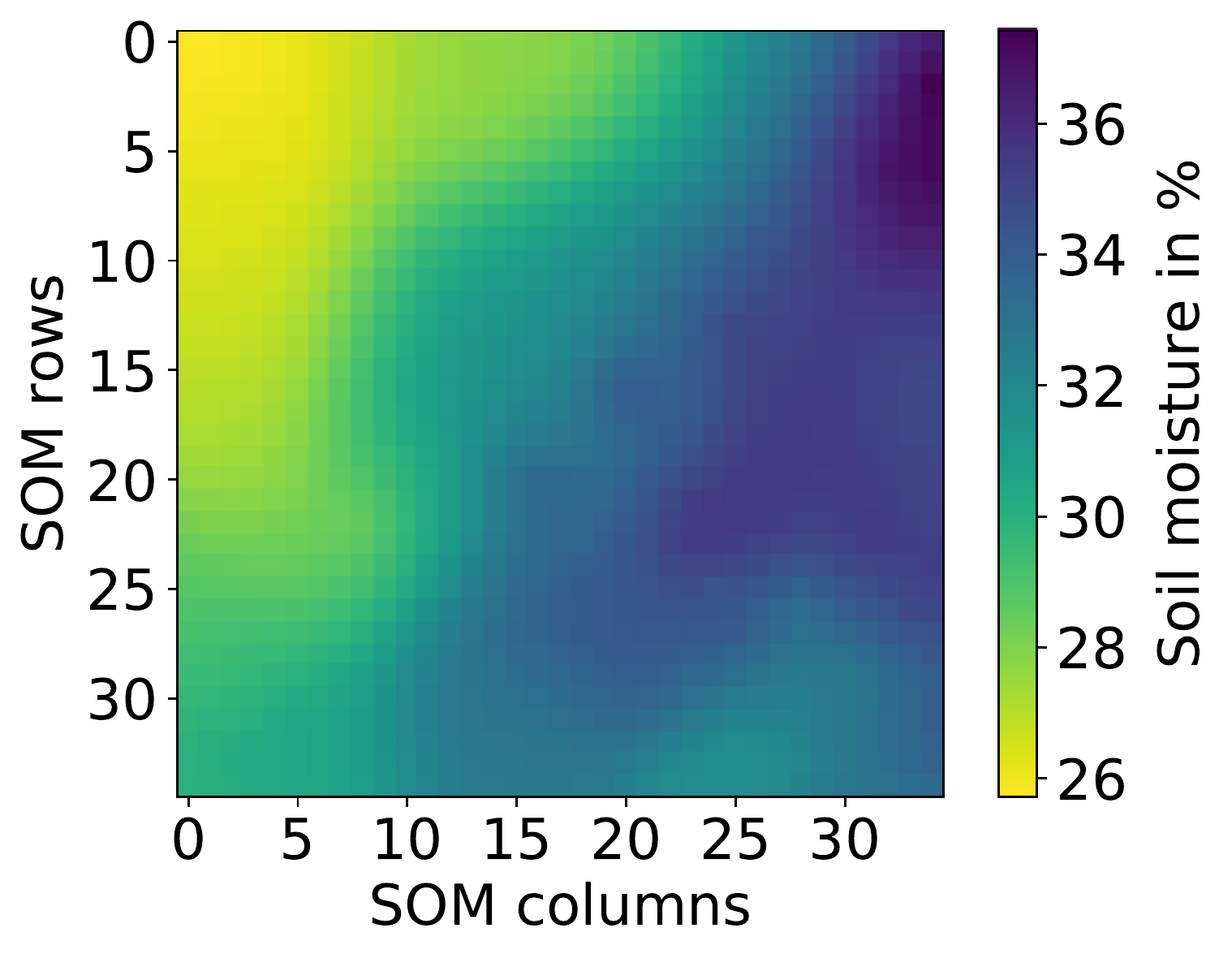}\label{fig:eval_reg_output}}
	\caption{Regression SOM distributions of (a)~the BMUs of the dataset and (b)~the regression output calculated for each node.\label{fig:eval_reg_all}}
\end{figure}

In \Cref{fig:eval_reg_histo}, the distribution of the BMUs of the soil-moisture dataset is shown.
No clear maximum exists, rather a random and uniform distribution is recognizable.
\Cref{fig:eval_reg_histo} illustrates further that despite the fact that the dataset is smaller than the number of SOM nodes, the training takes advantage of the whole SOM grid.
The spread over the whole SOM grid makes generalization possible.
The continuous regression output for each SOM node is presented in \Cref{fig:eval_reg_output}.
Although the SOM nodes outnumber the input datapoints, each SOM node is linked to a soil moisture value.

\subsection{Classification of land cover}
\label{sec:eval:sub:cla}
%

The Salinas valley dataset\footnote{\url{http://www.ehu.eus/ccwintco/index.php/Hyperspectral_Remote_Sensing_Scenes}.} is a freely available land cover dataset consisting of $512\times 217$ pixels collected by the 224-band AVIRIS sensor in California.
The spatial resolution of this dataset is \SI{3.7}{\meter}.
Of the 224 bands in the range of \SIrange{400}{2500}{\nano\meter}, the 20 water absorption bands are discarded, namely bands 108-112, 154-167, 224.
The dataset consists of \num{54129} datapoints with reference data of 16 classes including vegetation classes and bare soil.
Compared to the dataset used in the evaluation of the regression SOM in \Cref{sec:eval:sub:reg}, this dataset is considered as large dataset.
We apply a 5-fold cross-validation on this dataset for the evaluation of the classification SOM.
The evaluation results are the average results over all five cross-validation combinations.

Similar to \Cref{sec:eval:sub:reg}, the default SOM hyperparameters are used except for the grid size and the number of iterations.
The grid size of the SOM is $40\times 20$.
The unsupervised SOM is trained with \num{5000} iterations and the supervised SOM is trained with \num{20000} iterations.
The hyperparameters of the classification SOM can be further optimized.
The RF classifier is set up with \num{100} estimators and the scikit-learn default hyperparameters~\citep[cf.][]{scikit-learn}.
For the evaluation, we choose the metrics overall accuracy (OA), average accuracy (AA) and Cohen's kappa coefficient $\kappa$.
The OA is defined as the ratio between the number of correctly classified datapoints and the size of the dataset.
The AA is the sum of the recall of each class divided by the number of classes with the recall of a class being the number of correctly classified instances (datapoints) of that class, divided by the total number of instances of that class.
Cohen's kappa coefficient $\kappa$ is defined as
\begin{align}
    \kappa = \frac{\mathrm{OA} - \theta}{1 - \theta},\label{eq:kappa}
\end{align}
with the hypothetical probability of chance agreement $\theta$.

\begin{figure*}
	\centering
	\includegraphics[width=0.9\textwidth]{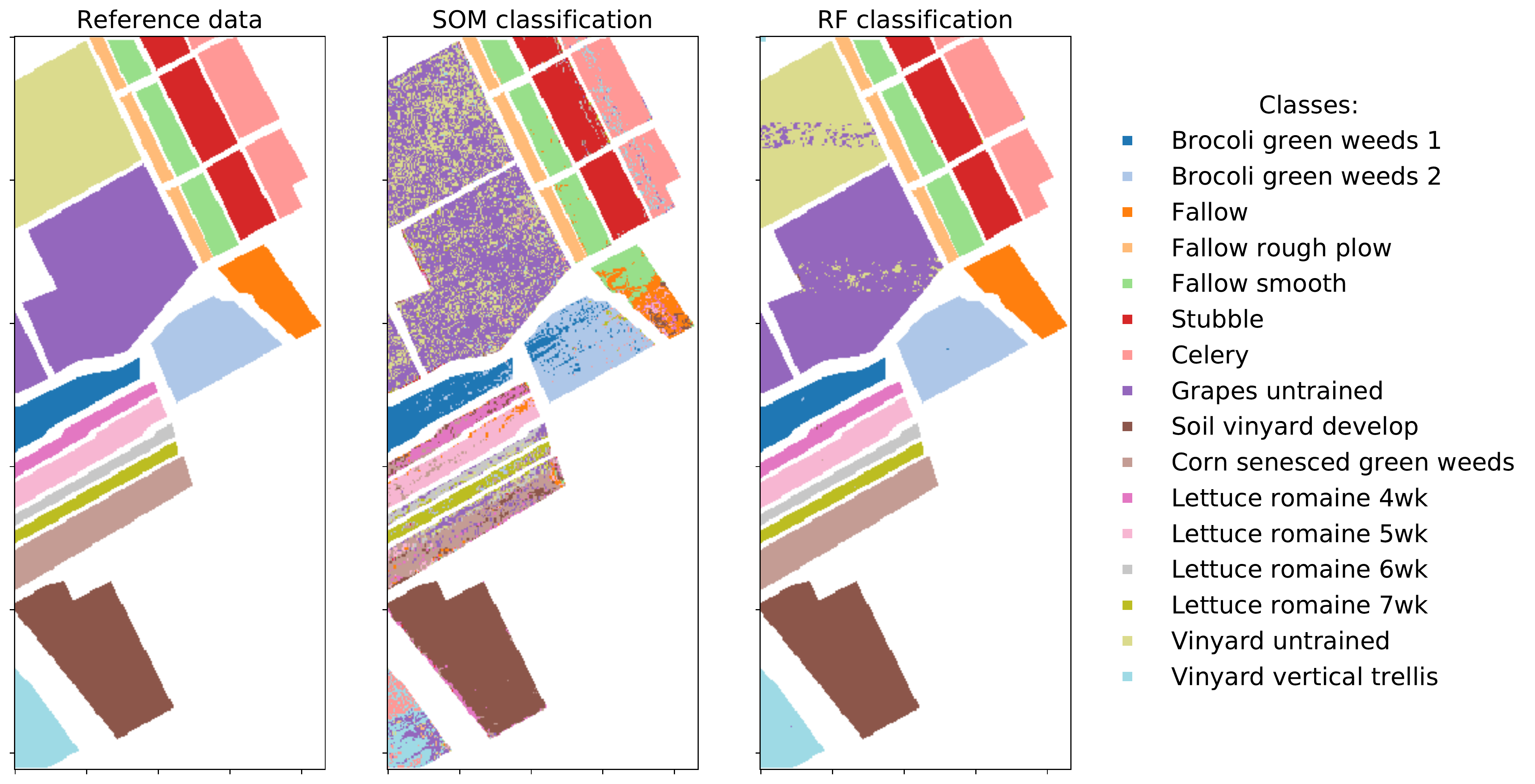}
	\caption{Map of the reference data (left) and the classification result of the classification SOM (center) and the RF classifier (right) on the Salinas Valley dataset. The white area is ignored.}
	\label{fig:eval_class_result}
\end{figure*}

The classification results of the complete dataset are shown in \Cref{fig:eval_class_result}.
The SOM achieves a test OA of $\SI{71.6}{\percent}$, AA $=\SI{72.8}{\percent}$ and  $\kappa = \SI{68.4}{\percent}$.
The training OA is $\SI{72.5}{\percent}$, the training AA is $\SI{73.7}{\percent}$ and the $\kappa$ score on the training subsets is $\SI{69.4}{\percent}$.
In contrast, the RF classifier achieves a test OA of $\SI{90.0}{\percent}$, AA $=\SI{93.8}{\percent}$ and $\kappa = \SI{88.9}{\percent}$ while the RF training metrics are all at \SI{100}{\percent}.
The RF classifier performs significantly better than the classification SOM which has not been fully optimized.
Analog to the regression (cf. \Cref{sec:eval:sub:reg}), the results for the classification SOM with respect to the training and test subsets are similar while the RF classifier shows overfitting on the training subset.

\begin{figure*}
	\centering
	\subfloat[]{\includegraphics[height=6cm, keepaspectratio]{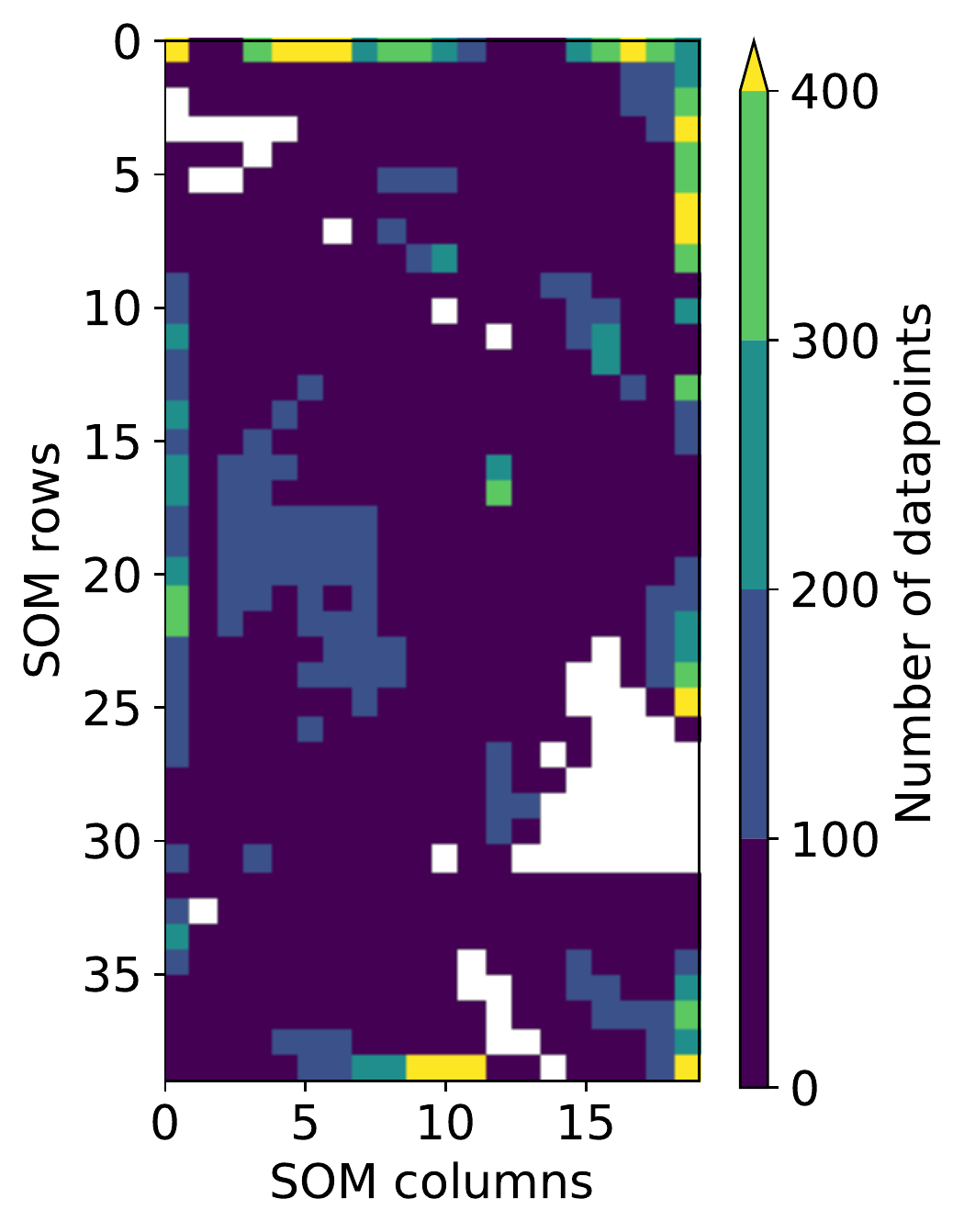}\label{fig:eval_class_histo}}\qquad\qquad
	\subfloat[]{\includegraphics[height=6cm, keepaspectratio]{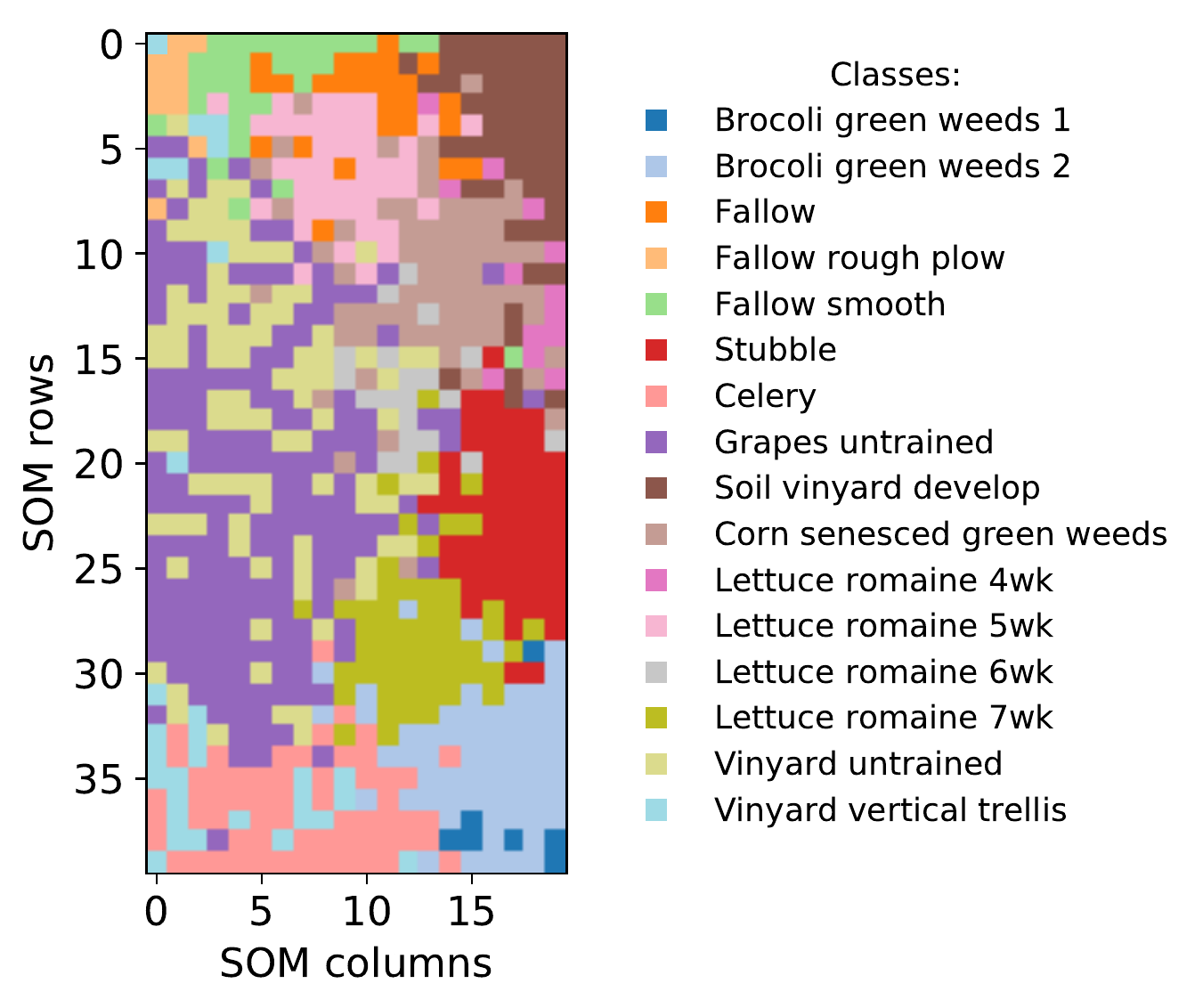}\label{fig:eval_class_output}}
	\caption{Classification SOM distributions of (a)~the BMUs of the dataset and of (b)~the classes linked to each node as output of the classification calculated.\label{fig:eval_class_all}}
\end{figure*}

In \Cref{fig:eval_class_histo}, the distribution of the BMUs of the dataset is illustrated.
Although the dataset is much larger compared to \Cref{fig:eval_reg_histo}, not all nodes are linked to a datapoint while some nodes are linked to a significant number of datapoints.
The distribution of the classification output of the SOM is shown in \Cref{fig:eval_class_output}.
Nodes assigned to the same classes are closer together on the SOM grid due to the inclusion of the neighborhood during the training process.

\subsection{Comparison of SuSi and other packages}
\label{sec:eval:sub:comp}

In the following section, we compare the SuSi framework with existing Software packages.
We compare it with the Python packages \textit{SOMPY}~\citep{moosavi2019sompy}, \textit{SimpSOM}~\citep{comitani2018simpsom}, \textit{MiniSom}~\citep{vettigli2019minisom}, \textit{TensorFlow SOM}~\citep{gorman2018amulti} and the \textsf{R} \textit{kohonen} package \citep{wehrens2018flexible}.
All entitled packages are freely available, regularly maintained (in the last year) and include unsupervised clustering.
\Cref{tab:comp} illustrates this comparison.
So far, no supervised SOM package for Python is available that matches the defined requirements requirements (cf. \Cref{tab:comp}).
The fact that the unsupervised SOM packages are all maintained regularly implies a significant interest in Python SOM packages.
Overall, the SuSi package is perfectly suited for an easy use and a variety of applications.

\begin{table*}
    \centering
    \caption{Comparison of the SuSi package with existing SOM packages. All packages are freely available and regularly maintained.}
    \begin{tabular}{lcccccc}
    \toprule
                                            & SuSi & SOMPY & SimpSOM & MiniSom & TensorFlow SOM & kohonen \\
    \midrule
         Simple (scikit-learn) syntax       & \checkmark & & \checkmark &  & & \\
         Comprehensive paper or documentation & \checkmark & &  & \checkmark &  & \\
         Well documented and structured code  & \checkmark & & \checkmark & \checkmark & \checkmark & \\
         \midrule
         Unsupervised clustering            & \checkmark & \checkmark & \checkmark & \checkmark & \checkmark & \checkmark \\
         Supervised regression              & \checkmark & &  &  &  & \checkmark\\
         Supervised classification          & \checkmark & &  &  &  & \checkmark\\
         \midrule
         Simple installation (e.g. Pypi)    & \checkmark &  & \checkmark & \checkmark &  & \checkmark\\
         GPU support                        &  &  &  & & \checkmark & \\
         Programming language               & Python 3 & Python 2 & Python 3 & Python 3 & Python 3 & \textsf{R}\\
    \bottomrule
    \end{tabular}
    \label{tab:comp}
\end{table*}

\section{Conclusion and outlook}
\label{sec:conclusion}

SOMs are applied in a variety of research areas.
In this paper, we introduce the \textbf{Su}pervised \textbf{S}elf-organ\textbf{i}zing maps (SuSi) package in Python.
It provides unsupervised and supervised SOM algorithms for free and easy usage.
The mathematical description of the package is presented in \Cref{sec:unsuper,sec:super}.
We demonstrate first regression and classification results in \Cref{sec:eval:sub:reg,sec:eval:sub:cla}.
Overall, the performance of the SuSi package is satisfactory, taking into account that a further optimization is possible.
The regression is performed on a small dataset with \num{679} datapoints while the classification SOM is applied on a large dataset with \num{54129} datapoints.
The application to these two datasets illustrates the ability of the SuSi framework to perform on differently sized datasets.
Although, the RF regressor and classifier perform better in the given tasks, the SOM performance metrics of the training and the test subset differ only slightly.
This shows the robustness of the SuSi framework.
Further, the performance metric based on the training dataset could function as \textit{out-of-bag estimate} for the dataset.
This implies that a dataset does not have to be split which improves training especially on small datasets.
Finally, we compare the SuSi framework against different existing SOM frameworks in Python and \textsf{R} with respect to e.g. features, documentation and availability.
We conclude that there is a significant interest in a standardized SOM package in Python which is perfectly covered by the SuSi framework.

In the future, the SuSi package will be extended, optimized and upgraded.
The handling of missing and incomplete data, as described in \cite{hagenbuchner2005a}, is one example for a possible new extension.
In addition, the 2D SOM grid visualizes the results of the SOM and therefore ensures to better understand the underlying dataset.
This ability to learn from underlying datasets can be extended as described e.g. by \cite{hsu2002selforganizing}.
Furthermore, we will present applications on new datasets as well as share best practices to make the SuSi framework as valuable as possible for its users.

{\footnotesize
	\begin{spacing}{0.9}		
	\bibliography{bibliography.bib}
	\end{spacing}
}

\end{document}